\documentclass[11pt,a4paper]{article}
\usepackage[hyperref]{eacl2021}
\usepackage{times}
\usepackage{latexsym}
\usepackage{url}
\usepackage{amsmath}
\usepackage{graphicx}
\usepackage{verbatim}
\usepackage{booktabs}
\usepackage{multirow}
\usepackage{makecell}

\renewcommand\vec[1]{\overrightarrow{#1}}
\newcommand\cev[1]{\overleftarrow{#1}}

\usepackage{microtype}

\aclfinalcopy 

\title{Canonical and Surface Morphological Segmentation for Nguni Languages}

\author{
Tumi Moeng \hspace{10pt}
Sheldon Reay \hspace{10pt}
Aaron Daniels \hspace{10pt}
Jan Buys \\
  Department of Computer Science \\
  University of Cape Town, South Africa \\
  \texttt{\{MNGTUM007,RYXSHE002,DNLAAR001\}@myuct.ac.za,jbuys@cs.uct.ac.za} 
}   

\date{}

\begin{document}
\maketitle
\begin{abstract}

Morphological Segmentation involves decomposing words into morphemes, the smallest meaning-bearing units of language. 
This is an important NLP task for morphologically-rich agglutinative languages such as the Southern African Nguni language group.
In this paper, we investigate supervised and unsupervised models for two variants of morphological segmentation: canonical and surface segmentation. 
We train sequence-to-sequence models for canonical segmentation, where the underlying morphemes may not be equal to the surface form of the word, 
and Conditional Random Fields (CRF) for surface segmentation.
Transformers outperform LSTMs with attention on canonical segmentation, obtaining an average F1 score of 72.5\% across 4 languages.
Feature-based CRFs outperform bidirectional LSTM-CRFs to obtain an average of 97.1\% F1 on surface segmentation. 
In the unsupervised setting, an entropy-based approach using a character-level LSTM language model fails to outperforms a Morfessor baseline, while on some of the languages neither approach perform much better than a random baseline. 
We hope that the high performance of the supervised segmentation models will help to facilitate the development of better NLP tools for Nguni languages. 
\end{abstract}

\section{Introduction}

Morphological Segmentation is the task of separating of words into their composite \emph{morphemes},
which are the smallest meaning-bearing units of a language \cite{creutz2007morph,ruokolainen-etal-2013-supervised}. This task is particularly important when applied to \emph{agglutinative} languages, which have words that are composed of aggregating morphemes, generally without making significant alterations to the spelling of the morphemes. 
Obtaining these morphemes enables analysis that can be applied to further Natural Language Processing (NLP) tasks \cite{creutz2007morph}. 
For example, breaking a word down to its composite morphemes before translation, or generating those morphemes one at a time, could lead to more accurate translation, especially in a low-resource scenario where limited training data is available.
Morphological analysis could also be used in the development of tools that could benefit language learners and assist linguists researching these languages.

\begin{figure}[t]
\centering
\captionsetup{labelfont=md}
\captionsetup{font=md}
\includegraphics[scale=0.28]{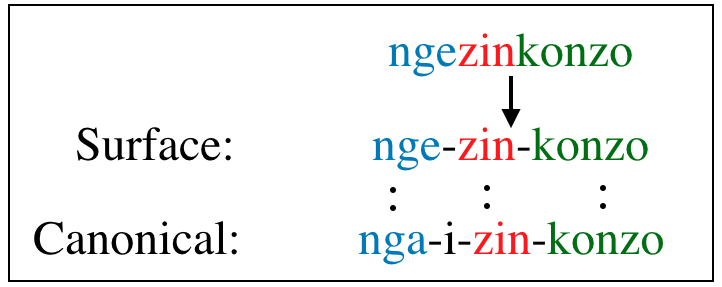} 
\caption {An isiZulu word with its canonical and surface segmentations.} 
\label{fig:first-example} 
\end{figure}

In this paper, we develop models for morphological segmentation for the Nguni languages, a group of low-resource Southern African languages.\footnote{The code of our models is available at \url{https://github.com/DarkPr0digy/MORPH_SEGMENT}}
We train supervised and unsupervised models for  isiNdebele, isiXhosa, isiZulu and siSwati, which are all official languages of South Africa.
Morphological segmentation is particularly applicable to Nguni languages because they are agglutinative and written conjunctively \cite{taljard2006comparison, bosch-pretorius-2017-computational}.
The only previous work we are aware of based on the datasets we use is a rule-based approach 
\cite{eiselen-puttkammer-2014-developing}, which 
our supervised models outperform substantially.  
\citet{cotterell-etal-2015-labeled} trained a semi-Markov CRF for isiZulu segmentation, but used a different corpus \cite{spiegler-etal-2010-ukwabelana}. 

We investigate supervised sequence-to-sequence models \cite{sutskever2014sequence, bahdanau2015neural, vaswani2017attention} as well as Conditional Random Fields \cite[CRFs;][]{lafferty-etal-2001-conditional,lample-etal-2016-neural,ma-hovy-2016-end} for segmentation. 
For sequence-to-sequence models we interpret the process of transforming a word into its segmented form as a character-level sequence transduction problem, which has previously been shown to be effective when applied to other languages \cite{wang-etal-2016-morphological,shao2017cross, ruzsics-samardzic-2017-neural}.
Sequence-to-sequence models are able to deal with input and output sequences of differing lengths, and subsequently to handle \emph{canonical} segmentation, where a morpheme may not be equal to the segment of the word that it corresponds as written \cite{kann-etal-2016-neural}.
The CRFs on the other hand are suitable for \emph{surface} segmentation, where the morphemes are a pure segmentation of the orthography of the word.
Figure \ref{fig:first-example} shows an example of a word with its canonical and surface segmentations. 

Canonical segmentation results show that the bidirectional LSTM with attention outperforms the LSTM without attention, while the Transformer leads to the best performance on all languages, with an average F1 of 72.5\%.
For surface segmentation, the feature-based CRF obtains an average F1 score of 97.1\% across the 4 languages, compared to 94.8\% for the Bi-LSTM CRF.

We also implemented an unsupervised entropy-based approach to morphological surface segmentation, based on character-based LSTM language models in the forward and backward directions.
We experimented with different entropy-based objective functions for segmentation, but none outperformed the Morfessor baseline. For some of the languages, neither approach perform much better than a random baseline.

\section{Background: Morphological Segmentation}

\subsection{Nguni Languages}

The Nguni language group consists of number of languages spoken in Southern Africa. isiZulu, isiXhosa, (Southern) isiNdebele and siSwati are all official languages of South Africa and constitute the majority of Nguni speakers (along with Zimbabwean Northern Ndebele). 
These languages are closely related to one another, with siSwati being a bit further apart from the rest, and Ndebele's vocabulary being influenced more by the neighbouring Sotho-Tswana languages.
All of these languages can be described as low-resource \cite{spiegler2008learning, bosch-etal-2008-experimental, bosch-pretorius-2017-computational}. 
The Nguni languages are agglutinative and are written conjunctively, meaning that words can be made up of many morphemes written unseparated \cite{spiegler2008learning, bosch-etal-2008-experimental}.\footnote{The Sotho-Tswana languages, the other major South African language group, are written \emph{disjunctively}: morphemes are generally written as separate words, despite the languages being agglutinative.} 
The meaning of a word is a function of all its morphemes. 
Therefore extracting the morphemes is essential for syntactic analysis and various forms of further text processing of these languages. 

Here is an example of agglutination in isiZulu \cite{bosch-pretorius-2017-computational}:
\begin{itemize}
  \item -phind-a - ``repeat''
  \item -phind-\textit{is}-a - ``cause to repeat''
  \item -phind-\textit{el}-\textit{el}-a - ``repeat again and again''
\end{itemize}

\subsection{Morphological Segmentation}

We distinguish two ways in which a word \emph{w} can be segmented, \emph{surface segmentation} and \emph{canonical segmentation} \cite{cotterell-etal-2016-joint}:
\begin{itemize}
    \item \textbf{Surface segmentation}: \emph{w} is segmented into a sequence of substrings, which when concatenated will result in \emph{w}.
    \item \textbf{Canonical segmentation}: \emph{w} is analyzed as a sequence of canonical morphemes representing the underlying forms of the morphemes, which may differ from their orthographic manifestation. 
\end{itemize}

The canonical segments correspond to the underlying morphemes used by linguists,
and may be more informative for downstream analysis than pure surface segmentation \cite{cotterell-etal-2016-joint}.

\subsection{Data}

The morphological annotations used in this paper come from the Annotated Text Corpora from the National Center for Human Technology \cite[NCHLT;][]{eiselen-puttkammer-2014-developing}.\footnote{Datasets are available at \url{https://repo.sadilar.org/handle/20.500.12185/7}} 
We use the isiNdebele, isiXhosa, isiZulu and siSwati corpora that are annotated with canonical morphological segmentations.
The morphemes are also labelled with their grammatical functions, but in this paper we will only consider the segmentation task and not the labelling task. 

The original annotation for the isiZulu word \emph{ngezinkonzo} is as follows, where the morpheme labels are given in square brackets: 
\begin{verbatim} 
  [RelConc]-nga[NPre]-i[NPrePre]
    -zin[BPre]-konzo[NStem]
\end{verbatim}

\begin{table}[t]
\begin{center}
\begin{tabular}{lccc}
\toprule
\textbf{Language} & \textbf{Train} & \textbf{Dev} & \textbf{Test} \\
 \midrule
 isiZulu  & 17 778&1 777& 3 298\\
 isiXhosa& 16 879&1 688& 3 004\\
 isiNdebele& 12 929&1 119& 2 553\\
 siSwati& 13 278&1 080& 1 347\\
 \bottomrule
\end{tabular}
\caption{Sizes of the morphological segmentation datasets (number of words) after preprocessing.}
\label{table:datasets}
\end{center}
\end{table}

The data consists of annotated running text.
We process the data to extract a set of annotated words.
We exclude punctuation, numbers, and words that are unsegmented in the annotations (as many of them are actually unannotated or are loan words).
The data is given with a training and test split. 
We ensure that there is no overlap between the training and test sets by removing all words appearing in both texts from the training data. 
This ensures that we are evaluating the ability of the models to generalize to unseen words. 
We split a development set from the training set in the same manner. 
The size of the processed datasets is given in Table \ref{table:datasets}.

\subsection{Generating Surface Segmentations}

We map the canonical segmentation annotations heuristically to corresponding surface segmentations.
We first check if the de-segmented canonical form is the same as the orthographic word, in which case the canonical and surface segmentations are equivalent. 
Otherwise, we compute the Levenshtein distance minimal edit operations from the de-segmented canonical form to the orthographic word. 
The operations are constrained so that each character in the input word corresponds to a character in the de-segmented canonical form. 
For example, in Figure \ref{fig:first-example}, the edit operations are to delete the single-character morpheme ``i'' and to replace the first ``a'' with ``e''. 

Finally, the sequence of edit operations is processed to align the canonical segments to the surface segments. This enables detecting the deletion of canonical segments, and mapping the morpheme boundaries in the canonical form to the orthography to create the surface segmentation. 

We computed a number of statistics to determine the efficacy of the method for obtaining surface segmentations. 
On average over all four languages, 45\% of the words' canonical segmentations differ from their surface segmentations.  
Of all the edit operations, 38.83\% of the operations were replacement operations, and the remaining 61.17\% were deletion operations. Finally, of all the segments generated in the surface form, 60.26\% of the segments are equal to the corresponding morpheme in the canonical form. 

\subsection{Evaluation}

The segmentation models are evaluated using precision, recall and F1 score of the morphemes identified for each word, compared to those in the annotated segmentation. 
We follow \citet{cotterell-etal-2016-joint} in treating the segmentation as a set of morphemes for evaluation purposes and computing the micro-F1 over the test set.
This contrast to the traditional approach to evaluating morphological segmentation with morpheme boundary identification accuracy. That method is not applicable to canonical segmentation, and we believe that basing the evaluation directly on morpheme identification is a better reflection of accuracy on this task.

\section{Canonical Segmentation with Sequence-to-Sequence Models}

We apply a Recurrent Neural Network (RNN) encoder-decoder model with attention \cite{bahdanau2015neural} as well as an encoder-decoder Transformer model \cite{vaswani2017attention} 
to the task of canonical morphological segmentation.
The task is formulated as transducing the given word's character sequence to an output character sequence consisting of the canonical form of the word together with the segment boundaries. 

\subsection{Bi-LSTM with Attention}

In an RNN-based encoder-decoder \cite{sutskever2014sequence} 
the encoder is an RNN which processes the input sequence \(x\) sequentially, updating the hidden
state of the RNN after reading each input element.
The decoder is another RNN which uses the final encoder hidden state as its initial hidden state to generate the output sequence. At each time step the previous output generated by the decoder is fed back into the RNN, which then produces the next output symbol, until a complete output sequence has been generated.
We use RNNs with Long-Short-Term-Memory (LSTM) cells \cite{hochreiter1997long} to avoid vanishing or exploding gradients.

The encoder can be extended to be bidirectional \cite{schuster1997bidirectional}, encoding the sequence using separate forward and backward LSTMs.
For each input element $x_j$, the encoder produces hidden layers $\vec{h_j}$ and $\cev{h_j}$ using a Bi-Directional RNN. The concatenation of these,
$h_j$, can be seen as a contextual representation of $x_j$.

To overcome limitations of using a single fixed context vector to encode the entire input sequence, an attention mechanism can be used, which enables the model to dynamically determine which parts of the input sequence to focus on as the sequence is traversed \cite{bahdanau2015neural}. 
The attention mechanism computes an alignment score $e_{ij}$ between the input at position $j$ and the output at position $i$, 
using a feed-forward network $a$:
\begin{equation}
    e_{ij} = a(s_{i-1} , h_j)
\end{equation}
where $s_{i-1}$ is the output from the previous decoder step and $h_j$ is the encoder representation at time step $j$.
The alignment scores are normalized with the softmax function, and the context vector $c_i$ at decoder time step $i$ is computed as a weighted average of the encoder representations $h_j$:
\begin{equation}
    c_{i} = \sum_{j=1}^{T_x} \mathrm{softmax}_j (e_{ij} ) h_j
\end{equation}
The decoder is therefore using the attention mechanism to dynamically calculate a context vector representation of the input sequence rather than using the (fixed) final encoder hidden state. 

\subsection{Transformer}

The Transformer is a sequence model which does not use recurrence and is instead based on the concept of self-attention \cite{vaswani2017attention}.
In the encoder-decoder framework, the transformer encoder uses multi-headed attention to calculate self-attention over the input sequence. 
The decoder uses two multi-headed attention blocks, over itself and over the encoder. 
Masking is used to prevent the first block from calculating self-attention over decoder positions which follow the current sequence position. 
A transformer block consists of alternating multi-headed self-attention and feed-forward layers.
Self-attention can be calculated in parallel, increasing computational efficiency. 
Due to the lack of recurrence, positional embeddings are used to encode the order of sequence elements in the encoder and the decoder.

\subsection{Experimental Setup}

The BiLSTM with attention was implemented by adapting an existing PyTorch code-base to support character-level transduction and a bidirectional encoder. 
The Transformer was implemented in the same code base.
As the model generates an output segmentation, it generates ``-'' to indicate a morpheme boundary. 
As a baseline we use an encoder-decoder without attention. Each model was trained on the dataset for each language independently. 

Based on hyperparameter tuning, the batch size was set at 32 or 64 for both the Transformer and the BiLSTM+Attention models. 
The best learning rates were 0.0001 for the biLSTM with attention, and 0.0005 for the Transformer.
The hidden dimension for both models was set at 256, and dropout was applied with a rate of 0.3. 
The biLSTM with attention used 2 layers. 
Adam \cite{kingma2017adam} was found to be a better optimizer than Stochastic Gradient Descent for both models. 

In order to compare with previous work, we ran the NCHLT rule-based segmenters \cite{eiselen-puttkammer-2014-developing}, which used the same (original) training corpus for model development, on our test set.\footnote{Available at \url{https://repo.sadilar.org/handle/20.500.12185/7/discover?filtertype=type&filter_relational_operator=equals&filter=Modules}}
The systems produce a set of multiple possible segmentations for some words; we compute an upper bound on performance by choosing the highest-scoring segmentation where multiple options are given.

\subsection{Results}

\begin{table}[t]
\centering
\begin{tabular}{lccc}
 \toprule
\textbf{Model}& \textbf{P}&\textbf{R}&\textbf{F1}\\
 \midrule
 \textbf{Rule-based} & 60.26	& 43.97& 	50.72 \\
 \midrule
isiZulu & 63.21 & 47.95 & 54.42 \\
isiXhosa & 59.85 & 48.88 & 53.81 \\
isiNdebele & 65.81 & 43.29 & 52.22 \\
siSwati & 52.16 & 35.76 & 42.43 \\
\midrule
\textbf{LSTM} & 64.78&56.92& 60.59\\
 \midrule
  isiZulu   & 68.72 & 60.25 & 64.2 \\
  isiXhosa   & 65.94&57.44&61.40\\
  isiNdebele   & 61.90&53.89& 57.61\\
  siSwati   & 62.54&56.10& 59.15\\
 \midrule
 \textbf{Bi-LSTM+Att}& 68.25&62.81& 65.41\\
 \midrule
  isiZulu   & 68.58&62.45& 65.37\\
  isiXhosa   & 70.06 &62.86& 66.26\\
  isiNdebele   & 64.90&59.67& 62.18\\
  siSwati   & 69.45&66.25 & 67.82 \\
 \midrule
 \textbf{Transformer}& 75.58&69.76& 72.54\\
 \midrule
  isiZulu   & 77.34 &71.04& 74.06\\
  isiXhosa   & 75.76&68.36& 71.87\\
  isiNdebele   & 73.14&66.67& 69.76\\
  siSwati  & 76.07& 72.96 & 74.48 \\
 \bottomrule
\end{tabular}
\caption{Canonical segmentation results. The Precision (P), Recall (R) and F1 Scores (F1) are given as percentages. The top line for each model reports the average over the 4 languages. For the rule-based systems we report the upper bound on accuracy among the possible segmentations produced.
} 
\label{table:canonical-results}
\end{table}

Table \ref{table:canonical-results} shows the canonical segmentation performance of each model on each language.
\citet{eiselen-puttkammer-2014-developing} reported F1 scores of 82\% to 85\% for the NCHLT rule-based model on the Nguni languages. However, they do not fully explain their experimental setup and if or how they disambiguate between multiple system outputs.
The rule-based systems perform substantially worse than any of our models, despite the fact that we are reporting an upper bound on its performance.
The gap is narrower on precision than on recall.
The rule-based system has higher precision than the LSTM-based model on isiNdebele, but everywhere else our models perform better. 
The rule-based systems produce canonical segmentations, but they produce segmentations matching the surface form for 92.7\% of words in the test set (averaged over the 4 languages), while in the gold annotations only 73.6\% have segmentations whose morphemes are equal to their surface forms. 

The Bi-LSTM+Attention model performs better than the encoder-decoder LSTM for all four languages, with an average increase in F1 Score by 4.82\%. 
siSwati, which had an average word length of 7 in the dataset (compared to 9 for the other three languages), achieved the highest increase out of all four languages with an increase of 8.67\%. 

\begin{table}[t]  
\begin{center}  
\begin{tabular}{lc}  
\toprule
\textbf{Model} & \textbf{Output} \\  
\midrule
Baseline& na-u-kuenza \\  
Bi-LSTM+Attention& na-u-ku-enza \\
Transformer& na-u-ku-enz-a \\  
\midrule
Target & \textbf{na-u-ku-enz-a} \\  
\bottomrule
\end{tabular}  
\caption {Sample model outputs for canonical segmentation, compared to the target (reference) segmentation.} 
\label{tab:seq-output-example}  
\end{center}  
\end{table}

The Transformer model showed the best results across all languages and metrics. It outperformed the baseline LSTM model by 11.95\%, and the Bi-LSTM+Attention model by 7.13\%. 
The F1 score for isiZulu improved by 9.7 percentage points from 65.37\% to 74.06\%. This shows the model's ability to learn features within the low resource language. 
The Transformer performed well within the low resource environment, and more data would likely improve its performance even further. 
The Transformer's performance was also the most consistent across languages: Its F1 scores have the lowest Standard Deviation among the three models. 
The Transformer results shows that relying solely on attention is a feasible approach for morphological segmentation. 
Table \ref{tab:seq-output-example} shows sample outputs from all three models compared to the target output.

\begin{figure}[t]
\centering
\captionsetup{labelfont=md}
\captionsetup{font=md}
\includegraphics[scale=0.3]{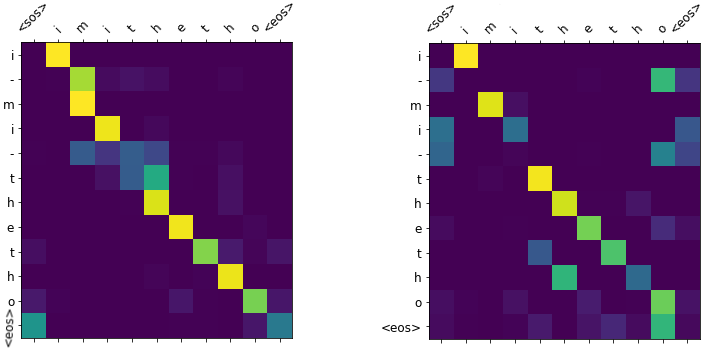}
\caption {Attention distributions of two of the Transformer's attention heads.} 
\label{fig:attention2} 
\end{figure}

Figure \ref{fig:attention2} shows two of the 8  encoder-to-decoder attention heads of the Transformer model.
The attention matrices show that the attention mechanism is able to align input segments to the correct corresponding output morphemes; this can be seen consistently across attention heads.

The results show that in a low-resource setting the models are able to learn and extract features from the languages at a satisfactory level. 
Nevertheless, the size of the datasets offers the best explanation of why the models didn't achieve higher accuracies. 
isiNdebele has the smallest training set, as well as the lowest performance across all the models.
siSwati also has a smaller dataset, but the words are shorter and consequently there are less segments, which makes the problem simpler.
Consequently, the attention-based models' highest performance is on the siSwati dataset. 
In contrast, the rule-based model and the baseline LSTM perform relatively worse on siSwati.

\section{Surface Segmentation with Conditional Random Fields}

\subsection{CRFs}

Conditional Random Fields (CRFs) are a class of discriminative, globally-normalized, probabilistic sequence labelling models \cite{lafferty-etal-2001-conditional}. 
We apply traditional feature-based (log-linear) CRFs as well as neural CRFs based on bidirectional LSTMs 
to the task of surface morphological segmentation.
A CRF takes a sequence $X$ as input and estimates the probability distribution $P(Y|X)$, where $Y$ is a sequence of the same length as $X$.
Every label $y_i$ in $Y$ corresponds to an input token $x_i$.

For morphological segmentation, the input is the word, represented as a character sequence, and the output is a label sequence encoding the segmentation. 
The possible labels are \emph{B}, \emph{M}, \emph{E}, and \emph{S}, representing, respectively, whether the character is the start of a new morpheme, a part of the current morpheme, the end of the current morpheme, or a single length morpheme. 

A CRF computes a feature score $s(X, y_i, y_{i+1})$ for each position $i$.
That is, it scores every possible assignment of a pair of labels in adjacent positions, so that the labels probabilities are correlated with each other and not independent.
The total unnormalized score that the CRF assigns to a given label sequence $Y$ is:
\begin{equation} 
  S \left( X, Y \right) = \sum_{i=0}^{n-1} s(X, y_i, y_{i+1})
\end{equation}

The probability can then be computed by normalizing the score over $Y^{|X|}$, the set of all possible label sequences of the same length as $X$:
\begin{equation} \label{eq6}
  p(Y|X)= \frac {e^{S(X,Y)}}
  {\sum_{\widetilde{Y} \in Y^{|X|}}
         e^{S(X,\widetilde{Y})}}
\end{equation}

The summation needs to be computed efficiently during training: This can be done using dynamic programming with the forward algorithm.
During testing, the model makes use of a related dynamic programming algorithm known as the Viterbi Algorithm to calculate the highest scoring label sequence for a given input.
The run time of the dynamic program is quadratic in the number of possible labels, but here that is a small set. 

\subsection{Bi-LSTM CRFs}

A traditional CRF is a log-linear model with a large number of manually-designed sparse, usually binary-valued, features. 
The model learns a weight for each feature. 
Alternatively, the scoring function can also be parameterized using neural networks, which can learn the features instead.
We use a bidirectional LSTM \cite{hochreiter1997long} to encode the input sequence. 
Our Bi-LSTM CRF follows that of previous work on neural sequence labelling \cite{lample-etal-2016-neural,ma-hovy-2016-end}.

To implement the CRF scoring function, the Bi-LSTM output is represented as an $n$ by $k$ matrix $P$, where $n$ is the number of words in the sequence and $k$ is the number of labels in the label alphabet. 
Each element $P_{ij}$ represents the (unnormalized) score of assigning label $y_i = j$ (as an index into the label vocabulary) to $x_i$.
The CRF score is defined as 
\begin{equation} 
  s \left( X, y_i, y_{i+1} \right) =  P_{i, y_{i}} + A_{y_{i},y_{i+1}}, 
\end{equation}
where $A$ is a learned square matrix of dimension $L+2$, where $L$ is the number of labels in the label alphabet, and the 2 added labels represent the start and the end of the sequence \cite{lample-etal-2016-neural}. 

\subsection{Implementation}

Our implementation of the feature-based CRF uses the 
\texttt{sklearn-crfsuite} library.\footnote{\url{https://github.com/TeamHG-Memex/sklearn-crfsuite}}
The features used are character $n$-grams with $n$ in the range 0 to 6, whether the character is a vowel or a consonant, and whether the character is uppercase or lowercase. 
We adapt an existing Bi-LSTM-CRF implementation suited to the segmentation task. 

For the feature-based CRF we tuned the choice of features, epsilon (which determines the convergence condition), and the number of training iterations.
The best hyperparameters were an episilon of 1e-7 and a maximum of 160 training iterations.

For the Bi-LSTM CRF we tuned the number of training epochs and the learning rate.
The best models were trained for 20 epochs, with a learning rate of 9e-4 for isiXhosa and isiZulu, and 4e-4 for isiNedebele and siSwati.  

\subsection{Results}

\begin{table*}[t]
\centering
  \begin{tabular}{lcccccc}
    \toprule 
    & \multicolumn{3}{c}{\textbf{Feature-based CRF}} & \multicolumn{3}{c}{\textbf{Bi-LSTM CRF}} \\
    \cmidrule(lr){2-4}\cmidrule(lr){5-7}
    \textbf{Language} & \textbf{P} & \textbf{R} & \textbf{F1} & \textbf{P} & \textbf{R} & \textbf{F1} \\
    \midrule 
    isiZulu & 97.88 & 96.82 & 97.35 & 96.64 & 96.64 & 96.64 \\
    isiXhosa & 97.16 & 97.13 & 97.14 & 94.88 & 95.61 & 95.24 \\
    isiNdebele & 97.94 & 96.62 & 97.27 & 96.59 & 96.21 & 96.40 \\
    siSwati & 97.17 & 96.40 & 96.78 & 90.59 & 91.48 & 91.03 \\
    \midrule 
    \emph{Average} & 97.54 &	96.74 &	97.14 &	94.68	& 94.99 &	94.83 \\
    \bottomrule
  \end{tabular}
  \caption{Results for the Surface Segmentation Task. 
  }
  \label{table:crf-1}
\end{table*}

Table \ref{table:crf-1} shows the results of the two CRF models on the task of surface segmentation.
The feature-based CRF yielded very high performance in the surface segmentation task with an average F1 score of 97.13\%. 
Surprisingly, the performance of the Bi-LSTM CRF is more than 3\% lower, with an average F1 score of 93.81\% across the four languages. 
The gap is substantially larger on SiSwati than on the other languages, with an F1 score of about 5\% lower. 
One potential reason for the lower performance of the BiLSTM CRF is the small size of the dataset.
In contrast to the sequence-to-sequence models, the performance drops on SiSwati rather than on isiNdebele, suggesting that it may be harder to tune the CRF on short sequences than the sequence-to-sequence models. 

While not directly comparable, 
\citet{cotterell-etal-2015-labeled} reported 90.16\% F1 for surface segmentation of isiZulu using a semi-Markov CRF on the Ukwabelana corpus. 
However, in addition to using a different corpus, they performed semi-supervised training using only 1 000 annotated training examples (together with a larger unannotated corpus).
For future work, we'd like to investigate further to what extend the performance gap between canonical and surface segmentation is due to the models compared to the greater inherent difficulty of the canonical segmentation task.

\section{Unsupervised Segmentation}

Unsupervised segmentation is important for low-resource languages as morphological annotations are unavailable for most of them.
We use Morfessor  \cite{creutz-lagus-2007-unsupervised}, a widely-used model for unsupervised segmentation, to benchmark unsupervised segmentation on these datasets, following the same preprocessing and evaluation setup as for supervised surface segmentation.
We use the Morfessor-Baseline model, which uses a segmentation optimization criteria based on Minimal Description Length. 
We compare this to a random segmentation baseline which inserts segment boundaries at random positions in a word, as a way to check whether the unsupervised models are learning anything useful.
We also implemented an entropy-based model, following previous work on unsupervised segmentation of isiXhosa \cite{mzamo2019evaluation,mzamo2019towards}.

\subsection{Entropy-based Model}

\citet{Shannon1948} introduced the concept of entropy, which measures the amount of information produced by an event or process.
The conditional entropy of $x_i$ in a sequence $x_{1:i}$ for a given probability model $P(x | x_{1:i-1})$ can be defined as 
\begin{equation}
    - \sum\limits_{\Tilde{x} \in x^*} P(\Tilde{x} | x_{1:i-1}) \log P(\Tilde{x} | x_{1:i-1}) 
\end{equation}
where $x^*$ is the set of possible values of sequence element $x$. 

The intuition behind entropy-based morphological segmentation is that inside a morpheme, each next character will be less surprising (so have a lower entropy) than the previous one, while at the start of a new morpheme the entropy will increase as the character is less predictable. 

Previous work used smoothed character n-gram language models to estimate the entropy  \cite{mzamo2019evaluation,mzamo2019towards}.
Here we use a character-level LSTM language model instead that encodes the entire word, learning to estimate the probabilities of successive characters. 
We trained language models in both forward (left-to-right) and backward directions. 
These models are used to obtain the left and right entropies of words, respectively.
We trained 2-layer LSTMs with hidden state size of 200, dropout rate 0.2, and SGD with an initial learning rate of 20 that is decreased during training.

We experimented with a number of different objective functions that use the entropies to decide where to segment a word. 
We found that the best strategy was to consider the sum of left and right entropies at each character position and to insert a segment boundary if the sum exceeds experimentally determined constants. 
We refer to this model as \emph{Constant Entropy}.
Constants 4, 3, 12 and 2.5 were used for isiNdebele, siSwati, isiXhosa and isiZulu, respectively.

We also experimented with inserting a segment boundary based on whether the entropy increases between adjacent positions, as well as an objective  that compares the sum of the left and right entropies to the mean over all the entropies in the word to perform relatively thresholding. 
These are similar to objective functions proposed by \citet{mzamo2019towards}.
However, in our experiments the constant entropy objective performed substantially better than either of those approaches.

\subsection{Results}

\begin{table}[t]
\centering
\begin{tabular}{lccc}
 \toprule
\textbf{Model}& \textbf{P}&\textbf{R}&\textbf{F1}\\
 \midrule
 \textbf{Random}&24.83 &14.76 &18.51\\
 \midrule
isiZulu & 24.15 & 14.97 & 18.48\\ 
isiXhosa & 23.23 & 14.32 & 17.72\\ 
isiNdebele & 25.91 & 15.27 & 19.22 \\ 
siSwati & 26.03 & 14.49 & 18.62\\ 
 \midrule
 \textbf{Morfessor}  &28.06 &27.57 &27.69\\
 \midrule
 isiZulu & 20.37 & 23.19 & 21.69\\ 
 isiXhosa & 27.21 & 29.04 & 28.10\\ 
 isiNdebele & 20.60 & 21.36 & 20.97\\ 
 siSwati & 44.05 & 36.67 & 40.02\\ 
 \midrule
\textbf{Constant Entropy}&30.49 &25.82 &25.73\\
 \midrule
isiZulu & 28.74 & 14.20 & 19.01\\ 
isiXhosa & 22.79 & 15.41 & 18.38\\ 
isiNdebele & 34.19 & 14.81 & 20.67\\ 
siSwati & 36.22 & 58.85 & 44.85\\ 
 \bottomrule
\end{tabular}
\caption{The Precision (P), Recall (R) and F1 scores for unsupervised morphological segmentation. 
} 
\label{table:unsupervised-results}
\end{table}

The results for the unsupervised models are given in Table \ref{table:unsupervised-results}.
Our entropy-based models do not outperform the Morfessor baseline.
The entropy-based approach outperforms the random baseline by 7.22\% on average, but was 1.96\%
lower than Morfessor. 

The results suggest that there are substantial structural differences between the languages.
In particular, both models perform substantially better on siSwati than on the other languages.
This could partly be explained by siSwati words having less segments on average than the other languages. 
The entropy-based model outperforms Morfessor on siSwati, with a particularly high recall of 58.8\%.
Both models perform only slightly above the random baseline for isiNdebele and isiZulu.
For isiXhosa, the entropy model has similarly low performance, while Morfessor does 10\% better, indicating that it was able to learn more structure of the language than our model.

\citet{mzamo2019towards,mzamo2019evaluation} also found that entropy-based models do not outperform Morfessor, although theirs were based on $n$-gram language models instead of LSTMs.
They reported scores of up to 77\% boundary identification accuracy. 
We obtained similar results when evaluating Morfessor using that metric, but on the morpheme-based F1 metric 
the results are much lower.

\section{Conclusions}

We developed supervised models for surface and canonical segmentation, as well as an unsupervised segmentation model, for 4 Nguni languages. 
Sequence-to-sequence models outperformed a rule-based baseline for canonical segmentation by a large margin, with Transformers obtaining the highest performance. 
The feature-based CRF obtained very high accuracies on the surface segmentation task, with an average F1 score of 97.1\%, outperforming the Bi-LSTM CRF. 

The strong supervised results opens new avenues to apply models for morphological segmentation to downstream language processing tasks for Nguni languages.
For future work, performance could possibly be improved further through semi-supervised training, as sequence-to-sequence models usually benefit when trained on larger datasets. 
Multilingual training could also be investigated, due to the similarities between the languages. 

The performance of all the unsupervised models are substantially lower than the supervised models.
We hypothesize that substantially different models will be required to make progress in unsupervised morphological segmentation.

\section*{Acknowledgments}

This work is based on research supported in part by the National Research Foundation of South Africa (Grant Number: 129850) 
and the South African Centre for High Performance Computing.
We thank Zola Mahlaza for valuable feedback, and Francois Meyer for running an additional baseline.

\bibliography{anthology,references}
\bibliographystyle{acl_natbib}

\end{document}